\documentclass{article}

\PassOptionsToPackage{numbers, compress}{natbib}

\usepackage[final]{bdl_2018}

\usepackage[utf8]{inputenc} 
\usepackage[T1]{fontenc}    
\usepackage{hyperref}       
\usepackage{url}            
\usepackage{booktabs}       
\usepackage{amsfonts}       
\usepackage{nicefrac}       
\usepackage{microtype}      

\usepackage{amsmath,amssymb,amsfonts}
\usepackage{algorithmic}
\usepackage{graphicx}
\usepackage{textcomp}
\usepackage{xcolor}
\usepackage{wrapfig,lipsum,booktabs}
\usepackage[cal=boondoxo,calscaled=1]{mathalfa}
\usepackage[rmdefault=ntxmi,sfdefault=iwona,scaled=1.0]{isomath}
\DeclareSymbolFont{letters}{OML}{ntxmi}{m}{it}
\def\BibTeX{{\rm B\kern-.05em{\sc i\kern-.025em b}\kern-.08em
    T\kern-.1667em\lower.7ex\hbox{E}\kern-.125emX}}

\title{Variational Self-attention Model for Sentence Representation}

\author{
  Qiang Zhang$^1$, Shangsong Liang$^2$, Emine Yilmaz$^1$ \\
  $^1\,$University College London, London, United Kingdom \\
  $^2\,$Sun Yat-sen University, Guangzhou, China \\
  \texttt{\{qiang.zhang.16, emine.yilmaz\}@ucl.ac.uk}, \texttt{liangshangsong@gmail.com} \\
}

\begin{document}

\maketitle

\begin{abstract}
This paper proposes a variational self-attention model (VSAM) that employs variational inference to derive self-attention. We model the self-attention vector as random variables by imposing a probabilistic distribution. The self-attention mechanism summarizes source information as an attention vector by a weighted sum, where the weights are a learned probabilistic distribution. Compared with conventional deterministic counterpart, the stochastic units incorporated by VSAM enable multi-modal distributions of attention.  Furthermore, by marginalizing over latent variables, VSAM is experimentally more robust against overfitting. 
Experiments on the stance detection task demonstrate the superiority of our method. 
\end{abstract}

\section{Background}
\subsection{Sentence representation}
A sentence usually consists of a sequence of discrete words or tokens $\boldsymbol{v} = [\boldsymbol{v}_1, \boldsymbol{v}_2, \dots, \boldsymbol{v}_n]$, where $\boldsymbol{v}_i$ can be a one-hot vector with the dimension $N$ equal to the number of unique tokens in the vocabulary. Pre-trained distributed word embeddings, such as Word2vec~\cite{DBLP:conf/nips/MikolovSCCD13} and GloVe~\cite{DBLP:conf/emnlp/PenningtonSM14}, have been developed to transform $\boldsymbol{v}_i$ into a lower-dimensional vector representation $\boldsymbol{x}_i$, whose dimension $D$ is much smaller than $N$. Thus, a sentence can be encoded in a more dense representation $\boldsymbol{x} = [\boldsymbol{x}_1, \boldsymbol{x}_2, \dots, \boldsymbol{x}_n]$. The encoding process can be written as: $\boldsymbol{x}=W^e\boldsymbol{v}$,  where $W^e$ is the 
transformation matrix. 
In the areas of natural language processing, the majority of deep learning methods (e.g. RNN and CNN) take $\boldsymbol{x}$ as the input and generate a compact vector representation $\boldsymbol{s}$ for a sentence: $\boldsymbol{s}=f^{\textrm{RNN}}(\boldsymbol{x})$, 
where $f^{\textrm{RNN}}(\cdot)$ indicates a RNN model. These methods consider the semantic dependencies between $\boldsymbol{x}_i$ and its context and hence believe that $\boldsymbol{s}$ summarizes the semantic information of the entire sentence. 

\vspace{-0.5em}
\subsection{Self-attention}
\vspace{-0.5em}
The attention mechanism~\cite{DBLP:journals/corr/BahdanauCB14,DBLP:conf/nips/VaswaniSPUJGKP17} has been proposed as an alignment score between elements from two vector representations. Specifically, given the vector representation of a query $\boldsymbol{q}$ and a token sequence $\boldsymbol{x} = [\boldsymbol{x}_1, \boldsymbol{x}_2, \dots, \boldsymbol{x}_n]$, the attention mechanism is to compute the alignment score between $\boldsymbol{x}_i$ and $\boldsymbol{q}$. 

Self-attention~\cite{DBLP:conf/emnlp/ParikhT0U16} is a special case of the attention mechanism, where $\boldsymbol{q}$ is replaced with a token embedding $\boldsymbol{x}_j$ from the input sequence itself. Self-attention is a method of encoding sequences of vectors by relating these vectors to each-other based on pairwise similarities. It measures the dependency between each pair of tokens, $\boldsymbol{x}_i$ and $\boldsymbol{x}_j$, from the same input sequence: $a_{i,j}=f^{\textrm{self-attention}}(\boldsymbol{x}_i, \boldsymbol{x}_j)$, 
where $f^{\textrm{self-attention}}(\cdot,\cdot)$ indicates a self-attention implementation.

Self-attention is very expressive and flexible for both long-term and local dependencies, which used to be respectively modeled by RNN and CNN. Moreover, the self-attention mechanism has fewer parameters and faster convergence than RNN. Recently, a variety of NLP tasks have experienced improvement brought by the self-attention mechanism.

\subsection{Neural variational inference}
Latent variable modeling is popular for many NLP tasks~\cite{DBLP:conf/icml/MiaoYB16,DBLP:conf/coling/BahuleyanMVP18}. It populates hidden representations to a region (in stead of a single point), making it possible to generate diversified data from the vector space or even control the generated samples. It is non-trivial to carry out effective and efficient inference for complex and deep models. Training neural networks as powerful function approximators through backpropagation has given rise to promising frameworks to latent variable modeling~\cite{DBLP:journals/corr/LinFSYXZB17,DBLP:conf/aaai/LuoZXW18}.

The modeling process builds a generative model and an inference model. A generative model
is to construct the joint distribution and somehow capture the dependencies between variables. The latent variable $\boldsymbol{z}$ the  can be seen as stochastic units in the generative model. The observed input and output nodes of $\boldsymbol{z}$ are denoted as $\boldsymbol{x}$ and $\boldsymbol{y}$ respectively. Hence the joint distribution of the generative model is:
\begin{equation}
p_\theta(\boldsymbol{x},\boldsymbol{y})= \int p_\theta(\boldsymbol{y}|\boldsymbol{z})p_\theta(\boldsymbol{z}|\boldsymbol{x})p(\boldsymbol{x})d\boldsymbol{z},
\end{equation} 
where $\theta$ parameters the generative distributions $p_\theta(\boldsymbol{y}|\boldsymbol{z})$ and $p_\theta(\boldsymbol{z}|\boldsymbol{x})$.
The variational lower bound is:
\begin{align}
\mathcal{L} & = \mathbb{E}_{q_{\phi}(\boldsymbol{z})}[\log (p_\theta(\boldsymbol{y}|\boldsymbol{z}) p_\theta(\boldsymbol{z}|\boldsymbol{x})
p(\boldsymbol{x})) - \log q_{\phi}(\boldsymbol{z})] = \int \log 
\frac{p_\theta(\boldsymbol{y}|\boldsymbol{z})
p_\theta(\boldsymbol{z}|\boldsymbol{x})
p(\boldsymbol{x})}{q_{\phi}(\boldsymbol{z})}  q_{\phi}(\boldsymbol{z})d\boldsymbol{z}  \displaybreak[3]\nonumber \\
& \leqslant \log\int 
p_\theta(\boldsymbol{y}|\boldsymbol{z})
p_\theta(\boldsymbol{z}|\boldsymbol{x})
p(\boldsymbol{x})  d\boldsymbol{z} = \log p_\theta(\boldsymbol{x},\boldsymbol{y})
\end{align}
In order to derive the evidence lower bound, the variational distribution $q_{\phi}(\boldsymbol{z})$ should approach the true posterior distribution $p_\theta(\boldsymbol{z}|\boldsymbol{x},\boldsymbol{y})$. A parametrized diagonal Gaussian distribution $\mathcal{N}(\boldsymbol{z}|\boldsymbol{\mu}(\boldsymbol{x},\boldsymbol{y}),\textrm{diag}(\boldsymbol{\sigma}^2(\boldsymbol{x},\boldsymbol{y})))$ is employed as $q_\phi(\boldsymbol{z}|\boldsymbol{x},\boldsymbol{y})$.

The inference model is to derive the variational distribution that approaches the posterior distribution of latent variables given observed variables.
The procedure to construct the inference model is:
\begin{enumerate}
\vspace{-0.3em}
\item Construct vector representations of the observed variables: $\boldsymbol{u}=f_x(\boldsymbol{x})$, $\boldsymbol{v}=f_y(\boldsymbol{y})$. \vspace{-0.3em}
\item Assemble a joint distribution: $\pi=g(\boldsymbol{u},\boldsymbol{v})$.\vspace{-0.3em}
\item Parameterize the variational distribution over the latent variables: $\boldsymbol{\mu}=l_1(\boldsymbol{\pi})$, $\log \boldsymbol{\sigma}=l_2(\boldsymbol{\pi})$.
\end{enumerate}
$f^x(\cdot)$ and $f^y(\cdot)$ are implemented as deep neural networks; $g(\cdot)$ can be an multi-layer perceptron that
fully connects the vector representations of the conditioning
variables; $l(\cdot)$ is a matrix transformation which computes the Gaussian distribution parameters. Sampling from
the variational distribution, i.e., $\boldsymbol{z} \thicksim  q_\phi(\boldsymbol{z}|\boldsymbol{x},\boldsymbol{y})$, allows to
conduct stochastic back-propagation to optimize the evidence lower
bound.

During training, the parameter $\theta$ of the generative model and the parameters $\phi$ of the inference model are updated by stochastic back-propagation based on samples $\boldsymbol{z}$ drawn from $q_\phi(\boldsymbol{z}|\boldsymbol{x},\boldsymbol{y})$. Let $L$ denote the total number of samples. The gradients w.r.t. $\theta$, are in the form of:
\begin{align}
\triangledown _{\theta} \mathcal{L} \simeq \frac{1}{L} \sum_{l=1}^{L} \triangledown _{\theta} \log 
(p_\theta(\boldsymbol{y}|\boldsymbol{z}^{(l)})
p_\theta(\boldsymbol{z}^{(l)}|\boldsymbol{x}))
\end{align} 

As about the gradients w.r.t. parameters $\phi$, we use the reparameterization trick $\boldsymbol{z}^{(l)}=\boldsymbol{\mu}+\boldsymbol{\sigma}\cdot \boldsymbol{\epsilon}^{(l)}$ and sample $\boldsymbol{\epsilon}^{(l)} \thicksim \mathcal{N}(0, \boldsymbol{I})$. $\phi$ can be updated by back-propagation of the gradients w.r.t. $\boldsymbol{\mu}$ and $\boldsymbol{\sigma}$:
\begin{align}
& \boldsymbol{\gamma}(\boldsymbol{z}) = \log (p_\theta(\boldsymbol{y}|\boldsymbol{z}) p_\theta(\boldsymbol{z}|\boldsymbol{x}))- \log q_\phi(\boldsymbol{z}|\boldsymbol{x},\boldsymbol{y}) \displaybreak[3]  \\
& \triangledown _{\boldsymbol{\mu}} \mathcal{L} \simeq \frac{1}{L} \sum_{l=1}^{L} \triangledown _{\boldsymbol{z}^{(l)}}  [\boldsymbol{\gamma}(\boldsymbol{z}^{(l)})], \triangledown _{\boldsymbol{\sigma}} \mathcal{L} \simeq \frac{1}{2L} \sum_{l=1}^{L} \boldsymbol{\epsilon}^{(l)} \triangledown _{\boldsymbol{z}^{(l)}}  [\boldsymbol{\gamma}(\boldsymbol{z}^{(l)})]
\end{align} 


\section{Variational Self-attention Model}
\begin{figure*}[ht]
	\centering
	\includegraphics[width=0.70\textwidth]{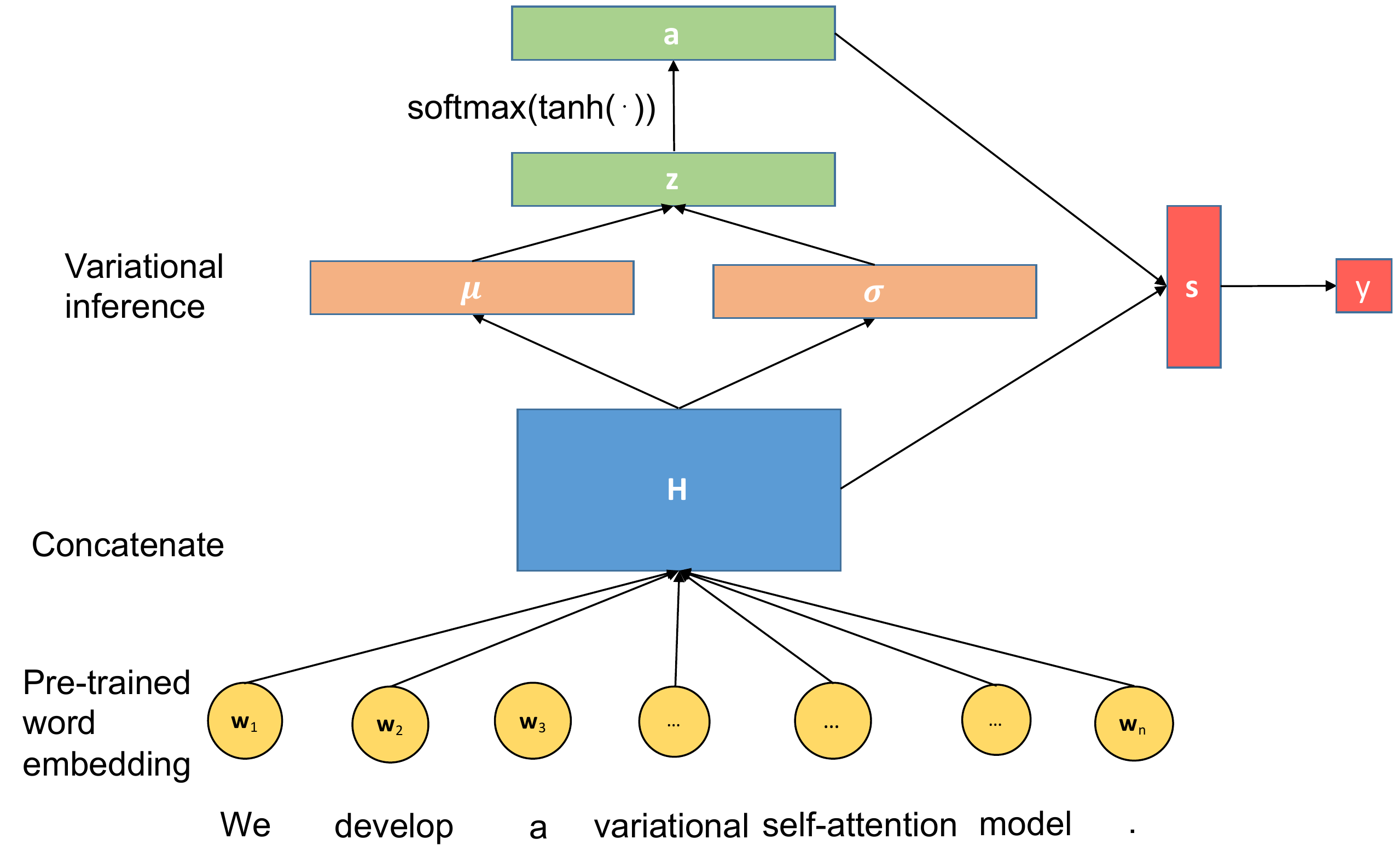}
	\vspace{-1em}
	\caption{The general framework of the variational self-attention model for sentence representation.}
	\centering
	\label{label:framework}
	\vspace{-1em}
\end{figure*}

In this paper we propose a Variational Self-attention Model (VSAM) that employs variational inference to learn self-attention. In doing so the model will implement a stochastic self-attention learning mechanism instead of the conventional deterministic one, and obtain a more salient inner-sentence semantic relationship. The framework of the model is shown in Figure $~\ref{label:framework}$.
Suppose we have a sentence $\boldsymbol{x} = [\boldsymbol{x}_1, \boldsymbol{x}_2, \dots, \boldsymbol{x}_n]$, where $\boldsymbol{x}_i$ is the pre-trained word embedding and $n$ is the number of words in the sentence. We concatenate the word embeddings to form a matrix $H \in \mathbb{R}^{D \times n}$, where $D$ is the dimension of the word embedding. 
We aim to learn semantic dependencies between every pair of tokens through self-attention. Instead of using the deterministic self-attention vector, VSAM employs a latent distribution $p_{\theta}(\boldsymbol{z}|H)$ to model semantic dependencies, which is a parameterized diagonal Gaussian $\mathcal{N}(\boldsymbol{z}|\boldsymbol{\mu}(H),\textrm{diag}(\boldsymbol{\sigma}^2(H)))$. Therefore, the self-attention model extracts an attention vector $\boldsymbol{a}$ based on the stochastic vector $\boldsymbol{z} \thicksim p_{\theta}(\boldsymbol{z}|H)$. 





The diagonal Gaussian conditional distribution $p_{\theta}(\boldsymbol{z}|H)$ can be calculated as follows:
\begin{align}
& \boldsymbol{\pi}_\theta = f_\theta(H) \displaybreak[3] \\
& \boldsymbol{\mu}_\theta = l_1(\boldsymbol{\pi}_\theta), \log \boldsymbol{\sigma}_\theta = l_2(\boldsymbol{\pi}_\theta) \displaybreak[3]\\
& p_{\theta}(\boldsymbol{z}|H) = \mathcal{N}(\boldsymbol{\mu}_\theta,\textrm{diag}(\boldsymbol{\sigma}_\theta^2)).
\end{align}
For each sentence embedding $H  $, the neural network generates the corresponding parameters $\boldsymbol{\mu}_\theta$ and $\boldsymbol{\sigma}_\theta$ that parametrize the latent self-attention distribution over the entire sentence semantics. 

The self-attention vector $\boldsymbol{a}\in \mathbb{R}^{n\times 1}$ can then be derived as: $\boldsymbol{a} = \mathrm{softmax}(\tanh(W^{z}\boldsymbol{z}))$. 
The final sentence vector representation $\boldsymbol{s}$ is the sentence embedding matrix $H$ weighted by the self-attention vector $\boldsymbol{a}$ as: $\boldsymbol{s} = H\boldsymbol{a}$, where $\boldsymbol{s}\in \mathbb{R}^{D \times 1}$. 
For the downstream application with expected output $\boldsymbol{y}$, the conditional probability distribution $p_\theta(\boldsymbol{y}|\boldsymbol{s})$ can be modeled as: $p_\theta(\boldsymbol{y}|\boldsymbol{s}) = g_\theta(\boldsymbol{s})$. 
As for the inference network, we follow the neural variational inference framework and construct a deep neural network as the inference network. We use $H$ and $\boldsymbol{y}$ to compute $q_{\phi}(\boldsymbol{z}|H,\boldsymbol{y})$ as: $\boldsymbol{\pi}_\phi = f_\phi(H,\boldsymbol{y})$. 
According to the joint representation $\boldsymbol{\pi}_\phi$, we can then generate the parameters $\boldsymbol{\mu}_\phi$ and $\boldsymbol{\sigma}_\phi$, which parameterize the variational distribution over the sentence semantics $\boldsymbol{z}$:

\begin{align}
& \boldsymbol{\mu}_\phi = l_3(\boldsymbol{\pi}_\phi), \log \boldsymbol{\sigma}_\phi = l_4(\boldsymbol{\pi}_\phi) \\
& q_{\phi}(\boldsymbol{z}|H,\boldsymbol{y}) = \mathcal{N}(\boldsymbol{\mu}_\phi,\textrm{diag}(\boldsymbol{\sigma}_\phi^2)).
\end{align}
To emphasize, although both $p_{\theta}(\boldsymbol{z}|H)$ and $q_{\phi}(\boldsymbol{z}|H,\boldsymbol{y})$ are modeled as parameterized Gaussian distributions, $q_{\phi}(\boldsymbol{z}|H,\boldsymbol{y})$ as an approximation only functions during inference by producing samples to compute the stochastic gradients, while $p_{\theta}(\boldsymbol{z}|H)$ is the generative distribution that generates the samples for predicting $\boldsymbol{y}$. 
To maximize the log-likelihood $\log p(\boldsymbol{y}|H)$ we use the variational lower bound. Based on the samples $\boldsymbol{z} \thicksim q_{\phi}(\boldsymbol{z}|H,\boldsymbol{y})$, the variational lower bound can be derived as
\begin{equation}
\begin{aligned}
\mathcal{L} & = \mathbb{E}_{q_{\phi}(\boldsymbol{z}|H,\boldsymbol{y})}[\log p_\theta(\boldsymbol{y}|H)] - D_{\mathrm{KL}}(q_{\phi}(\boldsymbol{z}|H,\boldsymbol{y})||p_{\theta}(\boldsymbol{z}|H))\\
& \leqslant \log\int p_\theta(\boldsymbol{y}|\boldsymbol{z}) p_{\theta}(\boldsymbol{z}|H) d\boldsymbol{z} = \log p(\boldsymbol{y}|H).
\end{aligned}
\end{equation}

The generative model parameters $\theta$ and the inference model parameters $\phi$ are updated jointly according to their stochastic gradients. In this case, $D_{\mathrm{KL}}(q_{\phi}(\boldsymbol{z}|H,\boldsymbol{y})||p_{\theta}(\boldsymbol{z}|H))$ can be analytically computed during the training process. 

\section{Experiments}
\begin{table}[t]
\small
\caption{Statistics of the FNC-1 dataset.}
\label{tab:dataset_statistics}
\centering
\begin{tabular}{@{}lcccc@{}}
\toprule
 Stance & \multicolumn{2}{c} {Training} & \multicolumn{2}{c}{Test}\\
&Number & Percentage&Number & Percentage\\
\toprule
 \emph{agree}       & {\color{white} 0}3,678  & {\color{white} 0}7.36  & 1,903 & {\color{white} 0}7.49  \\
 \emph{disagree}  & {\color{white} 00}840  & {\color{white} 0}1.68 &{\color{white} 0}697 & {\color{white} 0}2.74  \\
\emph{discuss}    & {\color{white} 0}8,909  &                        17.83 &  4,464 & 17.57   \\ 
 \emph{unrelated} &                        36,545  &           73.13 & 18,349 & 72.20   \\
\toprule
                   &                        49,972      &                  &                  25,413 &\\
\toprule
\end{tabular}
\end{table}
In this section, we describe our experimental setup. The task we address is to detect the stance of a piece of text towards a claim as one of the four classes: \emph{agree}, \emph{disagree}, \emph{discuss} and \emph{unrelated}~\cite{zhang2018ranking}. Experiments are conducted on the FNC-1 official dataset~\footnote{ \url{https://github.com/FakeNewsChallenge/fnc-1}}. The dataset are split into training and testing subsets, respectively; see Table~\ref{tab:dataset_statistics} for statistics of the split. We report classification accuracy and micro F1 metrics on test dataset for each type of stances.

Baselines for comparisons include: (1) \textbf{Average of Word2vec Embedding} refers to sentence embedding by averaging vectors of each word based on Word2vec. (2) \textbf{CNN-based Sentence Embedding} refers to sentence embedding by inputting the Word2vec embedding of each word to a convolutional neural network. (3) \textbf{Self-attention Sentence Embedding} refers to sentence embedding by calculating self-attention based sentence embedding, without variational inference. 

\begin{table}
\small
\caption{Performance comparison with the state-of-art algorithms on the FNC-1 test dataset.}
\label{tab:fnc}
\begin{center}
\vspace{-1em}
\begin{tabular}{lccccc}
\toprule
Model & \multicolumn{4}{c}{Accuracy (\%)} &Micro F1(\%) \\
                   & agree                        & disagree                       &   discuss          & unrelated\\
\hline
Average of Word2vec Embedding  & 12.43                         &  {\color{white} 0}1.30   &   43.32              & 74.24                  & 45.53\\
CNN-based Sentence Embedding   & 24.54                         &  {\color{white} 0}5.06   &   53.24   & 79.53                  & 81.72 \\ 
RNN-based Sentence Embedding     & 24.42                         &  {\color{white} 0}5.42   &  69.05               & 65.34                  & 78.70 \\
Self-attention Sentence Embedding             & 23.53                        &  {\color{white} 0}4.63   &   63.59               & 80.34                 & 80.11\\

Our model  &  28.53             &  10.43             &   65.43               & 82.43     & \textbf{83.54}\\
\bottomrule
\end{tabular}
\end{center}
\vspace{-2em}
\end{table}

Table~\ref{tab:fnc} shows a comparison of the detection performance. 
As for the \emph{micro F1} evaluation metric, our model achieves the highest performance (83.54\%) on the FNC-1 testing subset. 
The average method can lose emphasis or key word information in a claim; the CNN-based method can only capture local dependency among the text with limit to the filter size; the RNN-based method can obtain semantic relationship in a sequential manner. Differently, the self-attention method is able to combine embedding information between each pair of words, which means more accurate semantic matching of the claim and the piece of text. Compared with the deterministic self-attention, our method is a stochastic approach that is experimentally proven to better integrate each word embedding. 

\section{Conclusion}
We propose a variational self-attention model (VSAM) that builds a self-attention vector as random variables by imposing a probabilistic distribution. Compared with conventional deterministic counterpart, the stochastic units incorporated by VSAM allow multi-modal attention distributions. Furthermore, by marginalizing over the latent variables, VSAM is more robust against overfitting, which is important for small datasets. Experiments on the stance detection task demonstrate the superiority of our method.

\section*{Acknowledgments}
This project was funded by the EPSRC Fellowship titled "Task Based Information Retrieval", grant reference number EP/P024289/1.

\bibliographystyle{abbrv}

\end{document}